\documentclass[conference]{IEEEtran}
\IEEEoverridecommandlockouts
\usepackage{cite}
\usepackage[absolute]{textpos}
\usepackage{amsmath,amssymb,amsfonts}
\usepackage{algorithmic}
\usepackage{graphicx}
\usepackage{textcomp}
\usepackage{xcolor}
\usepackage[hidelinks]{hyperref}
\def\BibTeX{{\rm B\kern-.05em{\sc i\kern-.025em b}\kern-.08em
    T\kern-.1667em\lower.7ex\hbox{E}\kern-.125emX}}

\begin{document}

\title{Reinforcement Learning–Based Production Scheduling in an Industry-Based Coating Scenario Using the Digital Model Playground

\thanks{This work was supported by German Federal Ministry for Economic Affairs and Climate Action (BMWK) as part of the ‘‘Edge Data Economy Initiative’’ under Grant 01MD22001C.}
}

\author{\IEEEauthorblockN{1\textsuperscript{st} Arne Kröger}
\IEEEauthorblockA{\textit{Faculty of Management, } \\
\textit{Culture and Technology}\\
\textit{Osnabrück University of Applied Sciences}\\
Lingen, Germany \\
\href{https://orcid.org/0009-0001-4707-7238}{https://orcid.org/0009-0001-4707-7238}}
\and
\IEEEauthorblockN{2\textsuperscript{nd} Ralf Buschermöhle}
\IEEEauthorblockA{\textit{Faculty of Management, } \\
\textit{Culture and Technology}\\
\textit{Osnabrück University of Applied Sciences}\\
Lingen, Germany \\
\href{https://orcid.org/0009-0009-8560-3137}{https://orcid.org/0009-0009-8560-3137}}
\and
\IEEEauthorblockN{3\textsuperscript{rd} Wilhelm Hasselbring}
\IEEEauthorblockA{\textit{Department for Computer Science} \\
\textit{Software Engineering Group}\\
\textit{Kiel University}\\
Kiel, Germany \\
\href{https://orcid.org/0000-0001-6625-4335}{https://orcid.org/0000-0001-6625-4335}}
\and
\IEEEauthorblockN{4\textsuperscript{th} Henrik Wilbers}
\IEEEauthorblockA{\textit{Faculty of Management, } \\
\textit{Culture and Technology}\\
\textit{Osnabrück University of Applied Sciences}\\
Lingen, Germany}
}

\maketitle

\begin{textblock*}{8cm}(12.5cm, 1.0cm)
\footnotesize
\rightline{\textit{International Conference on Electrical, Computer and Energy Technologies (ICECET 2026)}} 
\rightline{\textit{6-9 July 2026, Rome-Italy}}
\end{textblock*}

\begin{textblock*}{8cm}(1.5cm, 26cm) 
\footnotesize
979-8-3195-0598-9/26/\$31.00 \textcopyright\,2026 IEEE
\end{textblock*}

\begin{abstract}
Production scheduling in complex manufacturing environments is challenging when sequence-dependent setup times, stochastic disturbances, and due-date constraints must be addressed simultaneously. While reinforcement learning (RL) methods have shown promising results in research, most studies rely on simplified benchmark processes, limiting their industrial relevance. This paper  demonstrates the applicability of RL-based scheduling in an industry-inspired coating process that reflects practical complexities such as sequence-dependent setup times, machine breakdowns, and variable utilization. The open-source Digital Model Playground (DMPG), a discrete event simulation framework, is used to model the scenario and to train RL agents. Two standard algorithms, Deep Q-Networks and Proximal Policy Optimization, are benchmarked against conventional dispatching rules to illustrate feasibility and to provide a transparent testbed for further research. Results indicate that RL-based scheduling achieves balanced improvements across key performance indicators, with PPO delivering the most robust performance. The main contribution of this work is to bridge the gap between academic research and industrial practice by validating RL-based scheduling in a realistic, shareable scenario and by providing a reusable open-source framework for future studies.
\end{abstract}

\begin{IEEEkeywords}
Production scheduling, reinforcement learning, Digital Model Playground, discrete event simulation, industry-inspired scenarios
\end{IEEEkeywords}

\section{Introduction}
The ongoing digitalization of production facilities enables the creation of digital twins (DTs), which can be used for production scheduling. This topic is of high industrial and academic interest. This is shown, among others, by Ouahabi et al., who identified that the number of publications in this area has tripled from 2019 to 2023 \cite{ouahabi_leveraging_2024}. However, they mention that most studies lack validation in real industrial environments. Another research trend in the field of production scheduling is reinforcement learning (RL). Modrak et al. showed a steep increase in publications in this field, from about 20 in 2018 to 200 in 2023 \cite{modrak_review_2024}. Nevertheless, similar to the use of DTs in production scheduling, there is a lack of studies validated in real-world scenarios of production scheduling, as stated by Panzer et al. \cite{panzer_deep_2021}. Our research goal is to combine these techniques and to create DTs from production facilities and use the simulation capability of the DT to train the RL agent for production scheduling. Since we assigned a Machine Learning Readiness Level (MLTRL) of 3 to RL- and simulation-based production scheduling \cite{seipolt_technology_2023}, our long-term goal is to develop the proof of concept, which would be the demonstration in a real-world scenario and marks the MLTRL 4 according to Lavin et al. \cite{lavin_technology_2022}. To provide a flexible framework to implement and optimize DTs from production facilities, we developed the Digital Model Playground (DMPG)~\cite{seipolt_reinforcement_2024}. The aim of this work is to show the feasibility of the DMPG and RL-based production scheduling in a complex real-world scenario, which is derived from a real use case. This process contains parallel machines, sequence-dependent setup times, machine breakdowns, and stochastic process times. We use the agent to minimize tardiness and setup times of the process. Furthermore, we will benchmark the trained agent with dispatching rules. 

The remainder of this paper is structured as follows: First, related work is discussed. Next, the experimental setup is shown in detail, as well as the approach to train the agent. We then show the result of the process and compare the agent to dispatching rules. The results are then discussed to contextualize them.

\section{Related Works}
Numerous studies have applied reinforcement learning (RL) to dynamic production scheduling, particularly for parallel or hybrid flow shop problems, with the main objective of reducing tardiness or setup times. Despite their differences in algorithms and environments, most approaches share two characteristics: (1) RL agents typically select dispatching rules rather than making direct scheduling decisions, and (2) experiments are conducted on simplified benchmark scenarios with limited industrial relevance.

Li et al.\ employed a two-stage RNN-based Proximal Policy Optimization (PPO) algorithm to minimize total tardiness in a parallel machine scheduling problem with due dates and family setups \cite{li_two-stage_2024}. Their method outperformed dispatching rules and metaheuristics, yet the model was restricted to a simplified parallel machine setting and did not explicitly aim to optimize setup times. 

Gerpott et al.\ integrated an Advantage Actor–Critic (A2C) algorithm into a two-stage hybrid flow shop to reduce tardiness and makespan \cite{gerpott_integration_2022}. The agent adaptively assigned predefined dispatching rules, but sequence-dependent setups and machine breakdowns were not considered, and setup time minimization was outside the scope. 

Wang et al.\ applied Deep Q-Learning (DQN) for hybrid flow shop scheduling, where the agent dynamically rescheduled jobs whenever machines became available \cite{wang_design_2024}. This adaptive use of dispatching rules outperformed static application of the same rules. A further study by Wang et al.\ extended this concept with an independent double DQN-based multi-agent approach in a two-stage hybrid flow shop with batch machines, reducing tardiness compared to the earliest due date (EDD) rule \cite{wang_independent_2022}. 

Qiu et al.\ proposed a multi-level action coupling deep Q-network to select dispatching rules in a flexible assembly flow shop \cite{qiu_multi-level_2024}. Their approach consistently outperformed EDD and standard DQN, highlighting the benefit of hierarchical action structures. Sun et al.\ introduced an improved PPO algorithm for hybrid blocking flow shops, reducing the action space by restricting the first scheduling steps to EDD and achieving superior performance compared to dispatching rules and metaheuristics \cite{sun_deep_2025}.

In summary, prior works consistently demonstrate the superiority of deep reinforcement learning (DRL) methods over dispatching rules and metaheuristics. The dominant strategy is to let RL agents adaptively select dispatching rules, with many studies focusing on algorithmic refinements of PPO or DQN. However, these contributions remain limited to generic benchmark problems, often neglecting real-world complexities such as sequence-dependent setup times, stochastic disturbances, and machine breakdowns. Moreover, existing studies rarely provide reusable frameworks, restricting the transfer of insights to practical applications. 

To address these shortcomings, our work builds upon the Digital Model Playground (DMPG) \cite{seipolt_reinforcement_2024}, an open-source discrete event simulation framework. By modeling a scenario derived from an actual industrial coating process, we validate RL scheduling in a realistic setting and benchmark it against dispatching rules, thus providing both methodological rigor and practical relevance. 

\section{Experimental Setup} \label{sec:setup}
This section describes the setup of the experiment. First, the production scenario is shown. Then, we describe the structure of the DMPG and how observation and action spaces are defined. Then, we show how the experiments are performed.

\subsection{Production Scenario}
\begin{figure*}
    \centering
    \includegraphics[width=1\linewidth]{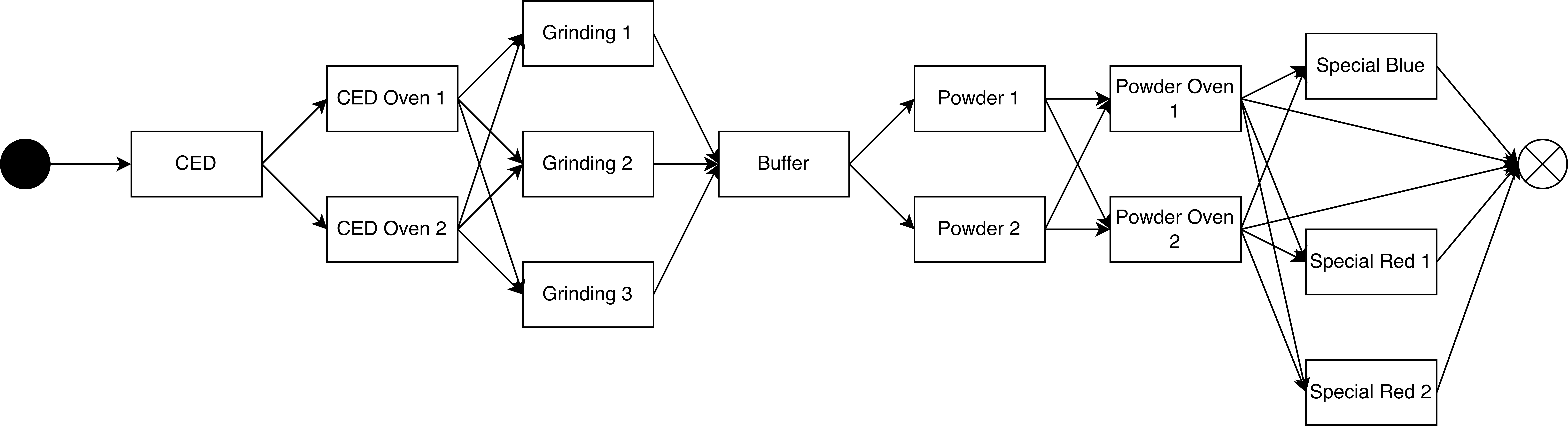}
    \caption{Simulated Coating Process.}
    \label{fig:ced-process}
\end{figure*}

The production scheduling investigated in this work is based on a cathodic electrodeposition (CED) process followed by powder coating. This two-step coating process is widely applied in various industries, including the automotive sector \cite{akafuah_evolution_2016}. The CED coating provides high corrosion resistance, while the subsequent powder coating enables customer-specific coloring and relatively short setup times when changing colors. The process is modeled as a discrete event simulation within the Digital Model Playground (DMPG), with the corresponding model available at \cite{digitaltwinml_dmpg_2026}. Fig. \ref{fig:ced-process} shows a flow sheet of the process. 

The first production step consists of CED coating, where products are cleaned, coated, and cured in one of two CED ovens. Any imperfections are repaired in one of three grinding stations. Subsequently, the products are buffered before being transferred to one of two powder cabins. In these powder stations, sequence-dependent setup times occur whenever the color is changed. After powder application, products are cured in one of two ovens. Depending on the color, an additional treatment step may follow: blue products require processing in a dedicated single station, red products in one of two parallel stations, while green and yellow products are finished directly after the curing step without further treatment.

The scheduling problem addresses two objectives: (1) minimizing setup times and associated costs and (2) minimizing deviations from planned finishing times. Changing colors in powder booths is particularly costly, as it requires thorough cleaning and leads to material losses. The cleaning effort depends on the color transition (e.g., switching between two yellow shades is less demanding than changing from blue to yellow). Consequently, the scheduling must optimize color sequences to reduce both setup time and cleaning-related costs. 

At the same time, products are expected to be completed at predefined due dates, which are often aligned with subsequent assembly processes. Minimizing deviations from these due dates reduces the need for buffer storage. While both earliness and lateness are penalized, delays are considered more severe than early completions and are therefore weighted accordingly in performance evaluation.

To evaluate process performance, three key indicators are defined:
\begin{itemize}
    \item \textbf{Setup Time:} The time required when switching between products of different colors, including chamber cleaning and refilling with a new color.
    \item \textbf{Weighted Deviation:} The deviation between actual and planned completion times. Delays are fully weighted, while early completions are weighted at half their value to reflect their lower impact.
    \item \textbf{Number Not Finished:} To prevent trivial solutions such as avoiding production altogether, the number of unfinished products is included as an additional criterion. 
\end{itemize}

The DES configuration is as follows: product arrivals are generated by four sources with random interarrival times. Each product is assigned a random due date. To analyze performance under varying utilization levels, the interarrival times are scaled by a utilization factor, randomly chosen between 0.7 and 1.3 at the beginning of each simulation run and kept constant thereafter. Therefore, the utilization is constant for every simulation, but the agent learns to perform under different conditions. All stations except the CED and powder stations are subject to random breakdowns, with failure intervals defined stochastically. Each station has a randomly drawn processing time. In case of multiple products waiting, the job with the earliest due date (EDD rule) is prioritized. The complete model is available at \cite{digitaltwinml_dmpg_2026}.

\subsection{Reinforcement Learning with the Digital Model Playground}
The Digital Model Playground (DMPG), introduced in December 2024, is a Python framework for discrete event simulation (DES) developed at Osnabrück University of Applied Sciences \cite{seipolt_reinforcement_2024}. Its goal is to provide an open-source alternative to commercial simulation tools. In addition to DES performance comparable to commercial software, the DMPG supports distributed execution of computationally intensive simulations, 2D and 3D visualization, and the integration of deep reinforcement learning (DRL) for process optimization. 

To optimize a production process within the DMPG, a simulation model must first be implemented. Decision-making is delegated to an object of the \textit{Connector} class, which links the simulation to the RL framework. Consequently, only two RL-specific components need to be implemented: the calculation of the reward and the construction of the state representation. In addition, a configuration file must be adapted to specify the hyperparameters. The framework automatically instantiates either a DQN or PPO agent using the TF-Agents library, the Reverb replay buffer, and a trainer–worker architecture. Experience collection can be performed in parallel by multiple worker processes. This allows users to fully exploit high-performance hardware without explicitly managing the complexity of parallelization. 

\begin{figure}
    \centering
    \includegraphics[width=0.5\linewidth]{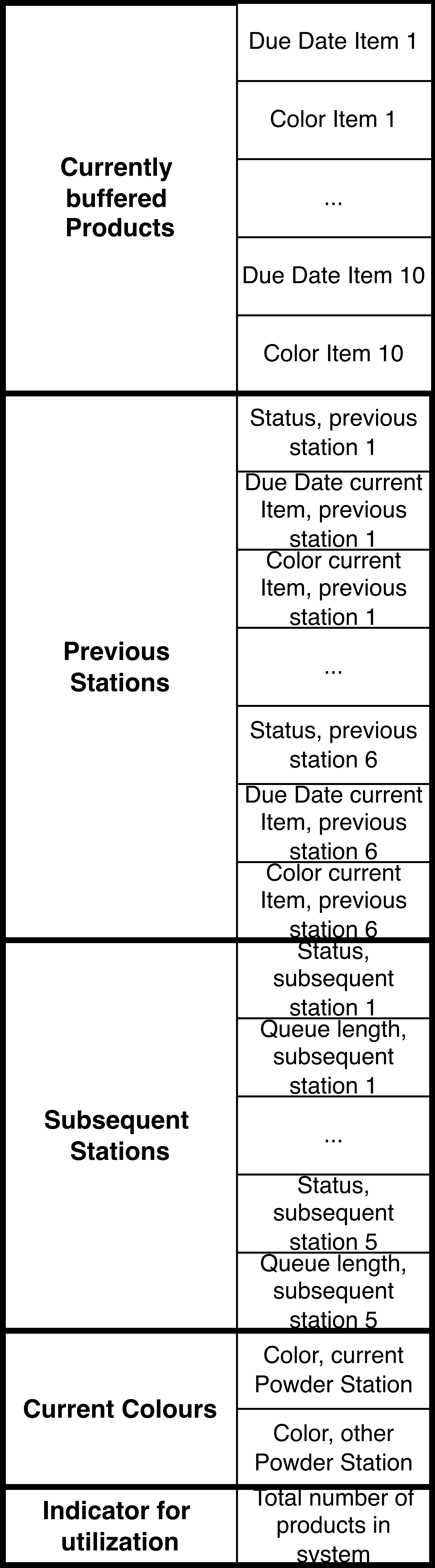}
    \caption{State representation of the production process used as input for the RL agent.}
    \label{fig:state}
\end{figure}

The state, which represents the process information on which the agent bases its actions, is shown in Fig.~\ref{fig:state}. It consists of five components: the currently buffered products, the status of previous and subsequent stations, the colors currently active in the powder stations, and an indicator of system utilization.  

The \textit{Currently Buffered Products} section includes both color and due date of the products waiting to be processed in the powder station. From previous and subsequent stations, the status is included to indicate whether a station is operational or down. For products being processed in previous stations, both color and due date are provided, while for subsequent stations only the queue length is considered. The state also encodes the current color configuration of the powder stations, enabling the agent to transfer experiences between Powder Station~1 and Powder Station~2. As an indicator of utilization, the total number of products present in the system is included. To distinguish colors, a one-hot encoding with four binary digits is applied. For example, yellow and blue are encoded by different positions of the active bit, while a chamber without color is represented by all zeros. With up to ten buffered products, the complete state vector comprises 121 positions.

Based on this state, the agent selects one out of five actions. Four actions correspond to the choice of a product color, in which case the product of that color with the earliest due date is scheduled. The fifth action represents idling, meaning no product is selected until a new one enters the buffer. 

The reward is derived from the outcome of the chosen action. It is primarily based on the performance parameters defined above. Instead of directly using the number of unfinished products, a delay metric is applied, defined as the time by which a product exceeds its due date plus an additional 3{,}000 time steps. This value is accumulated for all unfinished products, penalizing the agent for failing to complete jobs. To compute the final reward, all three performance measures are normalized and passed through a $\tanh$ activation function. Their weighted sum assigns the highest priority to minimizing delays (weight 5), followed by setup time (0.6) and weighted deviation (0.4). These values were selected heuristically to balance the trade-off between timeliness and setup efficiency. Given that product delays are critical, their corresponding weight is significantly the highest. While storing finished products in existing warehouses incurs costs, the expenses associated with cleaning powder chambers and the opportunity costs of products that could have been manufactured during setup time are substantially higher. Consequently, the weight assigned to setup time is prioritized accordingly.

Training is carried out on the high-performance cluster of Osnabrück University of Applied Sciences. Each job uses 80 processor threads, distributed across 60 worker processes for experience generation and 20 processes for transferring experiences to the training loop. In addition, one NVIDIA A100 GPU is allocated exclusively for training. Both PPO and DQN agents are trained, and the exact hyperparameter settings are provided in the corresponding configuration files of the public repository \cite{digitaltwinml_dmpg_2026}. 

To benchmark the RL agents, two conventional dispatching rules are applied: \textit{earliest due date (EDD)} and \textit{minimum setup time (MST)}. Since MST requires a certain buffer length, scheduling decisions are only made once a predefined number of products (0, 5, 10, 20, or 30) are waiting in the buffer. If fewer products are present, no scheduling decision is taken.

\section{Results}
\begin{figure}
    \centering
    \includegraphics[width=0.97\linewidth]{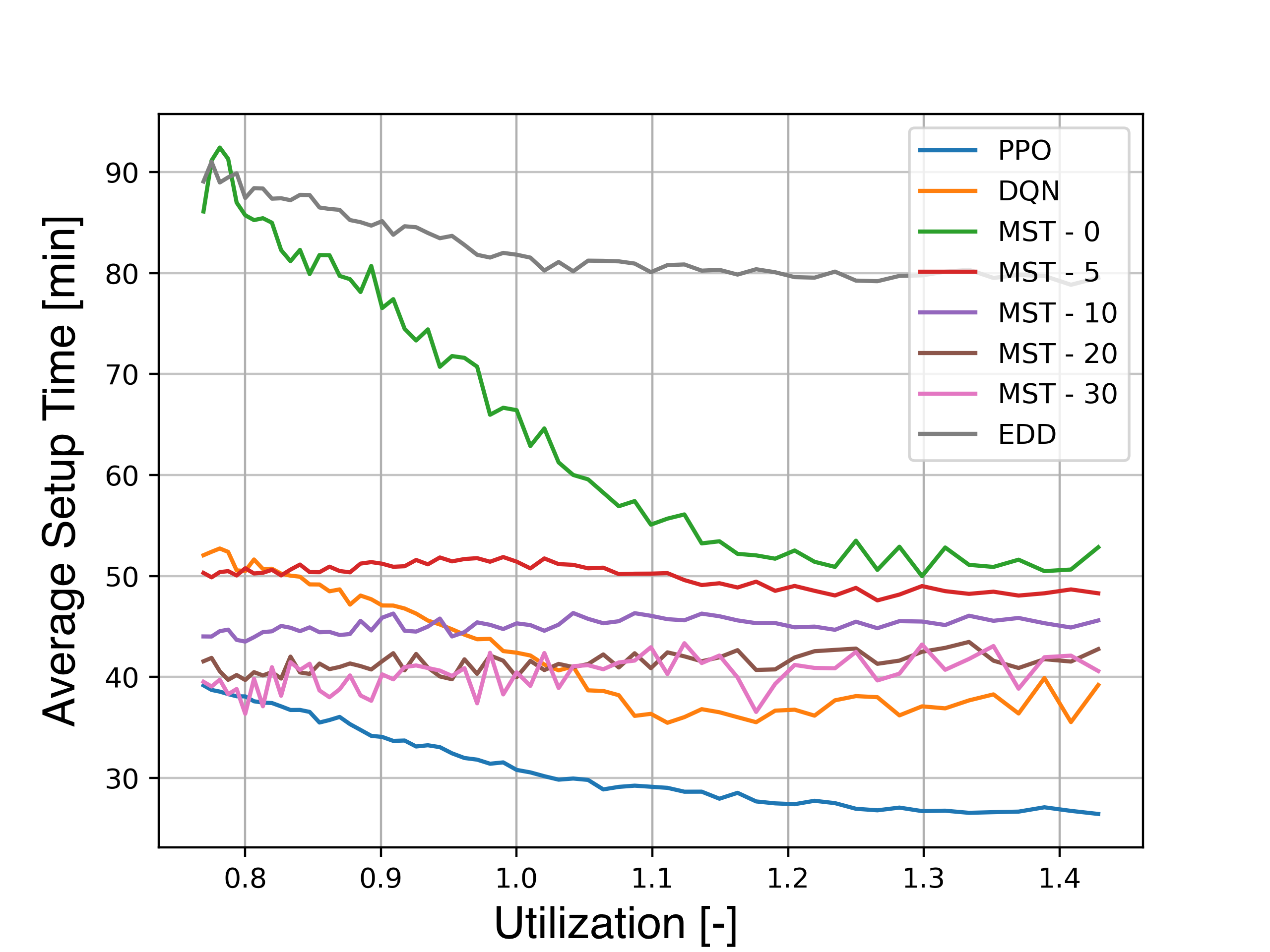}
    \caption{Comparison of the average setup time per product across scheduling methods.}
    \label{fig:setup}
\end{figure}

\begin{figure}
    \centering
    \includegraphics[width=0.97\linewidth]{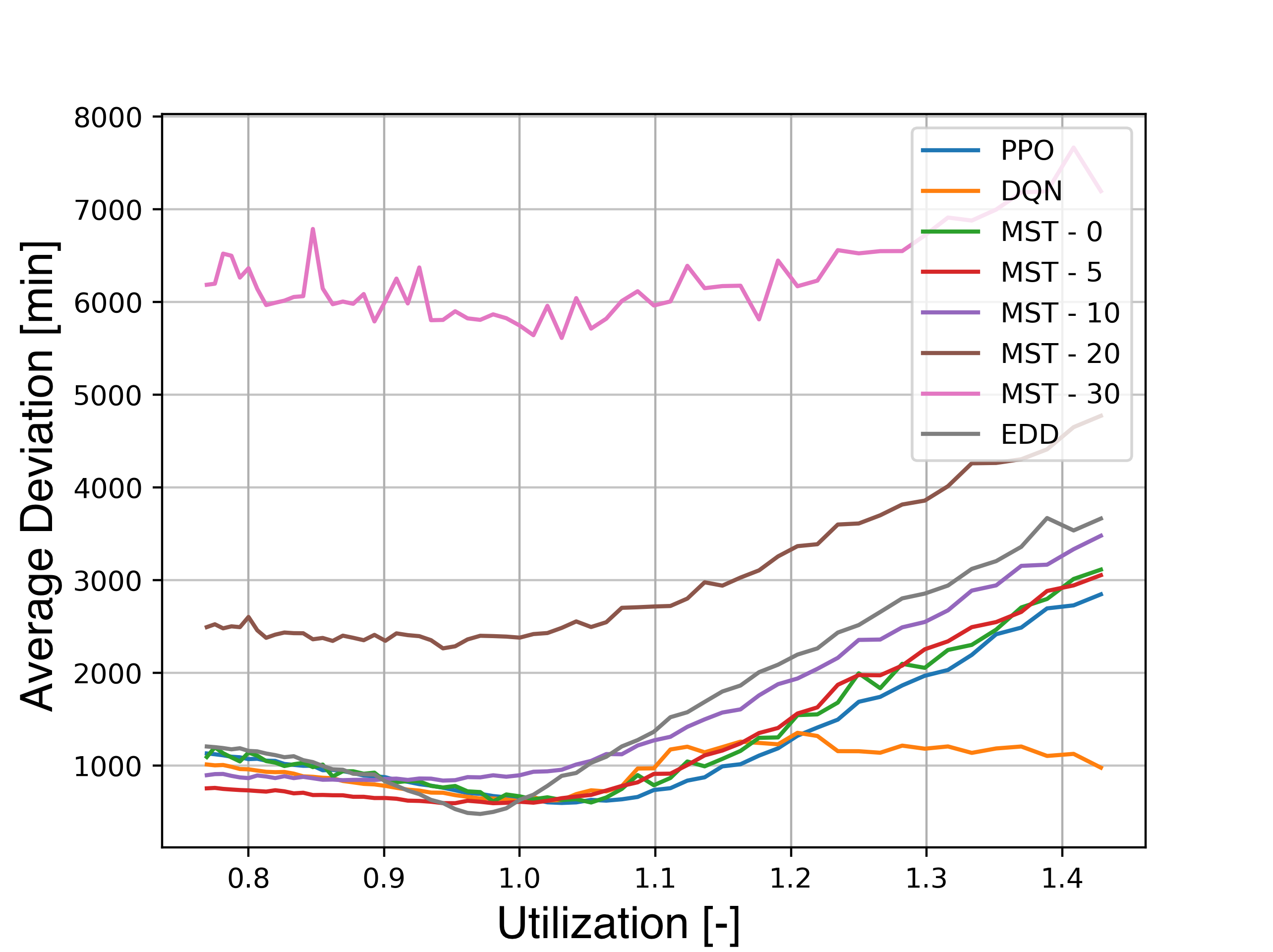}
    \caption{Comparison of the weighted deviation from the scheduled due date across scheduling methods.}
    \label{fig:deviation}
\end{figure}

\begin{figure}
    \centering
    \includegraphics[width=0.97\linewidth]{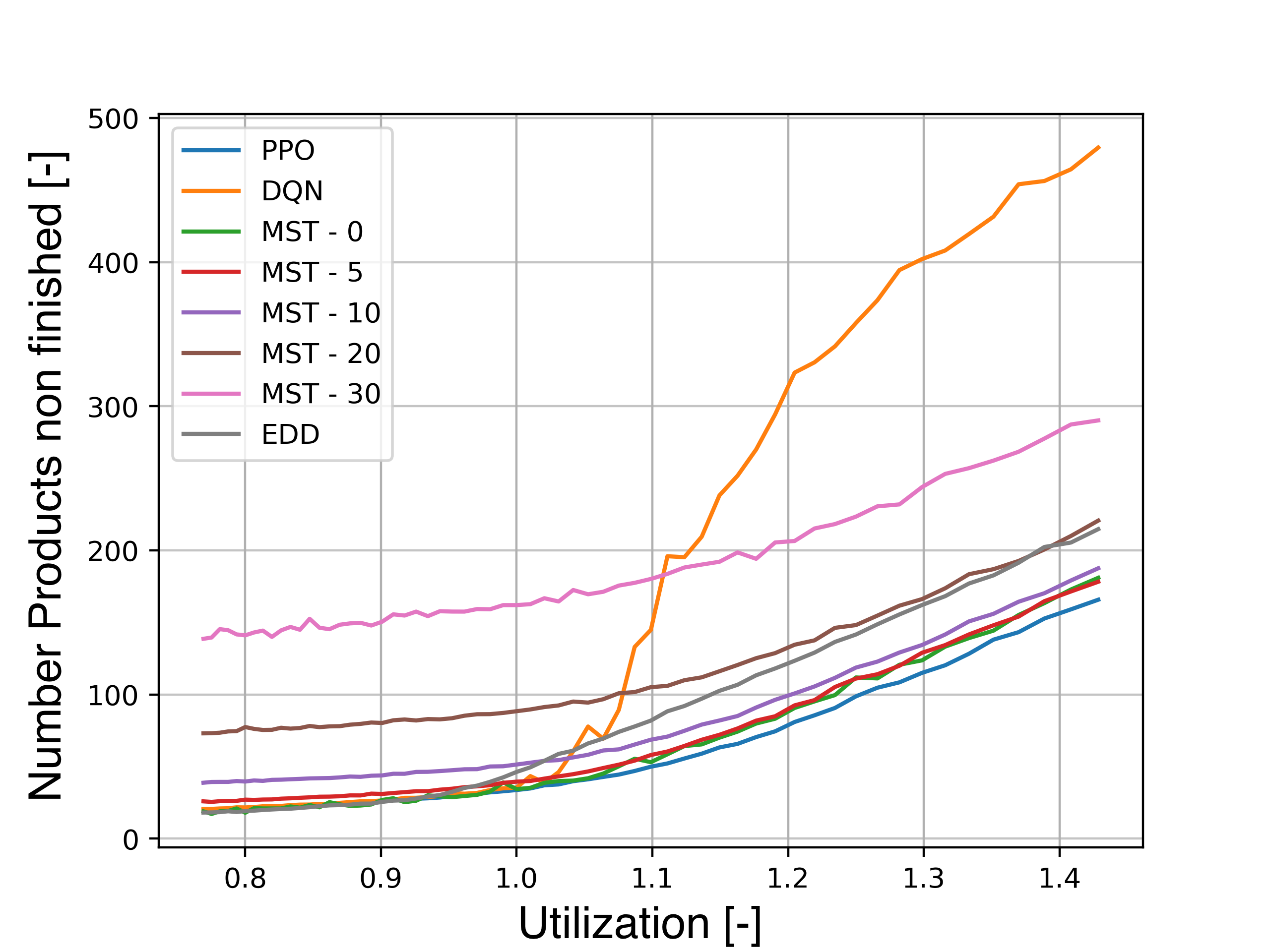}
    \caption{Comparison of the number of unfinished products across scheduling methods.}
    \label{fig:not_finished}
\end{figure}

This section compares the performance of the PPO and DQN agents with conventional dispatching rules. 
To analyze the behavior under varying process loads, the utilization factor is varied between 0.7 and 1.3, corresponding to utilizations from approximately 75\% to 140\%. 
For each utilization level, 100 simulation runs are conducted, and the average results are shown in Fig.~\ref{fig:setup}, Fig.~\ref{fig:deviation}, and Fig.~\ref{fig:not_finished}. 

Fig.~\ref{fig:setup} illustrates the average setup time per processed product. 
It is evident that the earliest due date (EDD) rule performs poorly with respect to setup time reduction. 
The minimum setup time rule (MST-0) and DQN show competitive performance at high utilization levels, but their setup times increase as utilization decreases. 
In contrast, PPO consistently outperforms all other approaches, particularly under high utilization.  

Fig.~\ref{fig:deviation} presents the weighted deviation from the scheduled due date. 
Across most utilization levels, the deviation remains relatively similar across algorithms, with the exception of MST rules using larger minimum buffer thresholds (MST-20 and MST-30), which perform substantially worse. 
At high utilization levels, DQN achieves the lowest deviation, but this advantage vanishes at lower utilizations. 
As Fig.~\ref{fig:not_finished} shows, this improvement is offset by a considerable increase in the number of unfinished products, highlighting a critical weakness of the trained DQN agent. 
Similarly, MST-20 and MST-30 result in significantly more unfinished products compared to the other methods.  

Table~\ref{tab:results} summarizes the average results across the entire utilization spectrum. 
The values confirm that larger minimum buffer lengths in MST reduce setup times but lead to considerably higher deviations and unfinished products. 
Overall, PPO achieves the best balance across all criteria, outperforming nearly all other algorithms. 
DQN yields the lowest weighted deviation on average but produces a substantially larger number of unfinished products. 
MST-5 achieves a slightly lower deviation than PPO but at the cost of approximately 60\% higher setup time.  

\begin{table}[]
    \centering
    \begin{tabular}{|c|c|c|c|}
    \hline
        \textbf{Algorithm} & \textbf{Setup Time} & \textbf{Deviation} & \textbf{Not Finished} \\
        \hline
         PPO     & 31.71 & 1,144 & 53.95 \\
         DQN     & 43.02 &   940 & 144.8 \\
         MST-0   & 67.05 & 1,210 & 57.48 \\
         MST-5   & 50.15 & 1,112 & 61.47 \\
         MST-10  & 44.85 & 1,390 & 72.67 \\
         MST-20  & 41.19 & 2,855 & 107.7 \\
         MST-30  & 40.20 & 6,283 & 180.2 \\
         EDD     & 83.17 & 1,489 & 72.94 \\
        \hline
    \end{tabular}
    \caption{Average setup time, weighted deviation, and number of unfinished products across all utilization levels.}
    \label{tab:results}
\end{table}

\section{Discussion}
In contrast to related work, our aim was not to optimize an abstract benchmark process but to derive a scenario from a real production facility. 
We modeled a two-step coating process combining CED and powder coating, as commonly found in the automotive industry. 
To increase realism, additional downstream steps were included for selected product groups (e.g., treatments analogous to labeling). 
The model further incorporates stochastic elements such as machine breakdowns and varying utilization levels. 
Consequently, it reflects several real-world factors. 
Nevertheless, some aspects were excluded for simplification, such as transportation times, resource capacities, or worker availability. 
Also, the restriction to four product colors is a simplification, and product arrivals are modeled as random, whereas in practice they follow at least partially planned schedules.  

We benchmarked DQN and PPO against conventional dispatching rules and showed that RL-based scheduling generally achieves superior performance. 
However, unlike our simulation model, real production facilities are supervised by human operators. 
For example, if the machine required for subsequent treatment of blue products fails, human controllers will proactively prevent powder stations from producing additional blue products. 
In contrast, dispatching rules cannot adapt dynamically and may continue producing blue products until the buffer is depleted. 
RL scheduling has the potential to react to such disruptions, but further studies are required to evaluate whether it can match or surpass human decision-making.  

Despite the simplifications, our findings support the applicability of RL-based scheduling to real-world processes. 
Still, performance must be compared against the combined decision-making of dispatching rules and human operators, which currently governs industrial practice.  

Several factors influence scheduling performance, and further research is required to identify processes where RL scheduling is particularly effective. 
For example, we did not investigate the effect of heterogeneous processing times or different numbers of parallel stations. 
In other scenarios, RL scheduling might therefore perform better or worse relative to dispatching rules.  

Since RL scheduling requires a trained agent, hyperparameters play a critical role. 
Minor adjustments can significantly alter agent performance. 
As we did not conduct a hyperparameter optimization study, the chosen parameters may not be optimal. 
It is therefore possible that alternative hyperparameter configurations could lead to a DQN agent outperforming PPO.  

Moreover, alternative approaches to production scheduling exist, such as evolutionary algorithms, metaheuristics, or dynamic programming. 
Further comparative studies are needed to establish guidelines for practitioners to select the most suitable scheduling approach for specific industrial settings.  

An important observation concerns the trade-off observed for MST rules between setup time and deviation. 
Larger buffer thresholds reduce setup times by enabling more efficient color sequencing, but at the same time lead to significantly higher deviations and a larger number of unfinished products. 
This illustrates the inherent conflict between efficiency and due-date adherence when using conventional dispatching rules. 
In contrast, RL-based scheduling does not exhibit this trade-off to the same extent: PPO achieves competitive results across all performance criteria without one metric degrading massively in favor of another. 
This balanced performance underlines the potential of RL methods to overcome classical scheduling conflicts that are difficult to address with rule-based approaches.

\section{Conclusion}
This work presented a reinforcement learning–based approach to production scheduling in an industry-derived coating process scenario. 
Unlike prior studies relying on highly idealized benchmarks, the modeled scenario incorporates multiple real-world complexities, such as sequence-dependent setup times, stochastic disturbances, machine breakdowns, and utilization variability. 
By benchmarking PPO and DQN against classical dispatching rules, we demonstrated that RL agents achieve competitive or superior results, with PPO consistently delivering the most robust performance. 
In particular, RL scheduling overcomes the classical trade-off observed in dispatching rules — such as the conflict between minimizing setup times and adhering to due dates — by balancing all performance measures without one aspect deteriorating significantly.  

By implementing the scenario within the open-source Digital Model Playground (DMPG), we provide a reusable and transparent testbed for both researchers and practitioners. 
This strengthens reproducibility and fosters comparative studies on scheduling algorithms in realistic industrial settings.  

Nevertheless, several challenges remain. 
RL scheduling performance is sensitive to hyperparameter selection, and the scalability of our approach to other industrial processes is yet to be validated. 
Moreover, the interaction of RL scheduling with human expertise—currently a key factor in real-world operations—has not been investigated.  

As future work, we aim to benchmark RL-based scheduling not only against dispatching rules but also against human decision-making in real production environments. 
This will provide a decisive proof of concept and shed light on the practical advantages and limitations of RL scheduling. 
Furthermore, extending the DMPG with more advanced RL algorithms or alternative optimization techniques (e.g., evolutionary methods or metaheuristics) will open opportunities to identify suitable approaches for different industrial contexts.  

\providecommand{\doi}[1]{DOI: \href{https://doi.org/#1}{#1}}

\end{document}